\documentclass{OAGM}
\OAGMarXiv{1304.1876}

\usepackage[utf8]{inputenc}
\usepackage{multicol}
\usepackage{longtable}

\usepackage{amsmath,amssymb,amsthm}

\usepackage{multirow}

\usepackage{graphicx}
\usepackage{subfigure}

\usepackage{algorithm}
\usepackage{algorithmic}

\renewcommand{\v}[1]{\mathbf{#1}}
\DeclareMathOperator*{\argmin}{arg\,min}

\title{Filament and Flare Detection in H$\alpha$~image~sequences}

\author{Gernot Riegler$^\dag$, Thomas Pock$^\dag$, Werner Pötzi$^\P$, Astrid Veronig$^\P$\\
  $^\dag$Institute for Computer Graphics and Vision, Graz University of Technology \\
  $^\P$Institute of Physics, University of Graz
  }

\begin{document}
\maketitle

\begin{abstract}
  Solar storms can have a major impact on the infrastructure of the earth. Some of the causing events are observable from ground in the H$\alpha$ spectral line. In this paper we propose a new method for the simultaneous detection of flares and filaments in H$\alpha$ image sequences. Therefore we perform several preprocessing steps to enhance and normalize the images. Based on the intensity values we segment the image by a variational approach. In a final postprecessing step we derive essential properties to classify the events and further demonstrate the performance by comparing our obtained results to the data annotated by an expert. The information produced by our method can be used for near real-time alerts and the statistical analysis of existing data by solar physicists. 
\end{abstract}

\section{Introduction}
The activity of the sun and the related space weather have gained wide interest in the last years. A recently published report by the Royal Academy of Engineering \cite{CannonFrang;AnglingMatthew;BarclayLes;CurryChales;Edwards2013} states the drastic impacts a large solar storm can have on the infrastructure of the earth and near-earth space. This includes the disturbance of the electrical power grid, pipelines and railway networks, satellites and navigation systems.

The cause of such solar storms are mainly solar flares and coronal mass ejections (\emph{CME}). Solar flares are an abrupt and enormous release of energy in a very short time. In a relative small area of the sun a single solar flare can create an explosion equivalent to several billion hydrogen bombs \cite{bhatnagar2005fundamentals}. CMEs on the other hand are correlated with filament eruptions \cite{lin2001prominence}. Both, flares and filament eruptions can be observed from ground based observatories in the H$\alpha$ spectral line, which is filtered by a narrow bandpass filter mounted on the telescope to create H$\alpha$ images.

\begin{figure}
	\center
	\subfigure[08:15:16]{
	\includegraphics[clip=true, trim = 1000px 650px 200px 600px, width=0.3\linewidth]{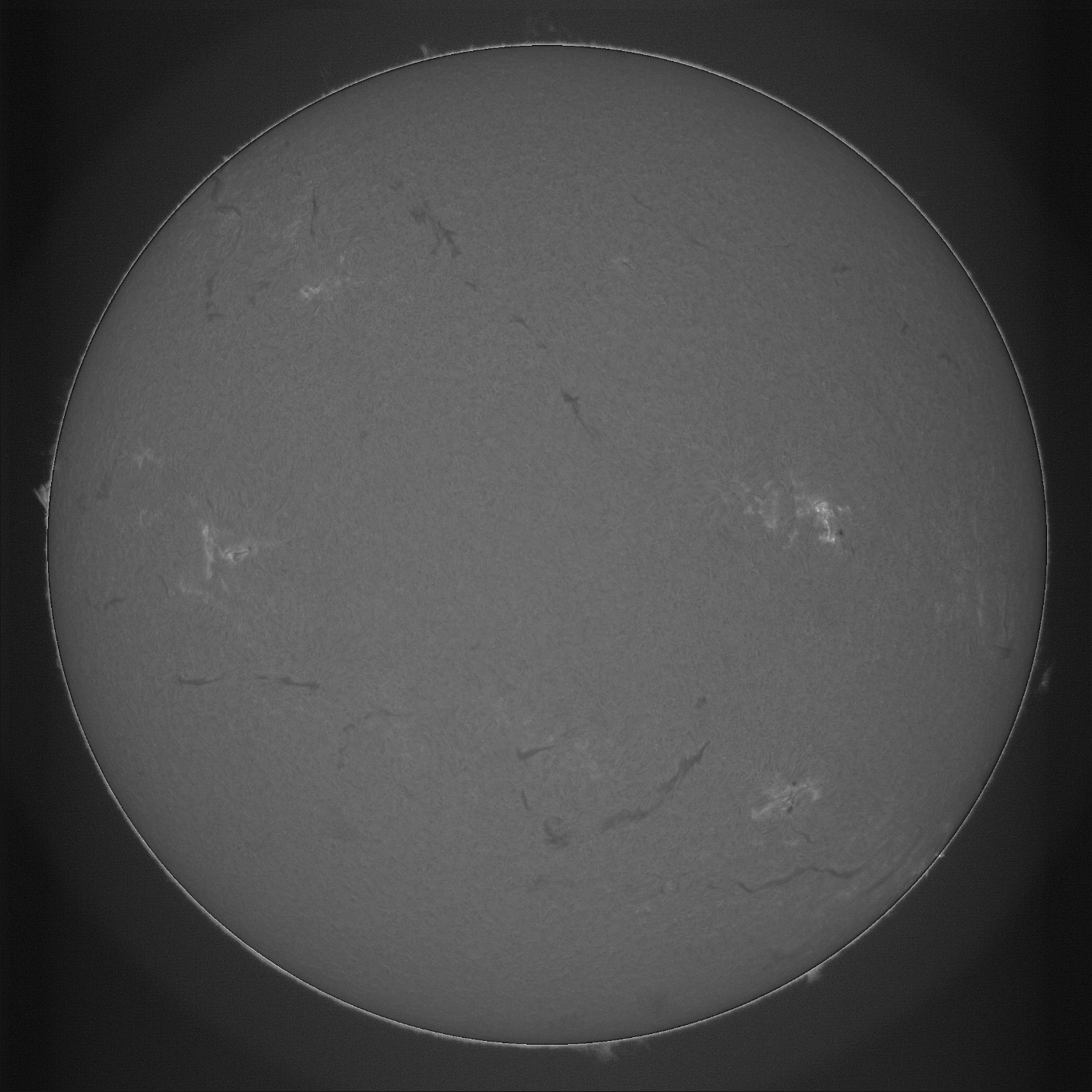}
	}
	\subfigure[08:20:09]{
	\includegraphics[clip=true, trim = 1000px 650px 200px 600px, width=0.3\linewidth]{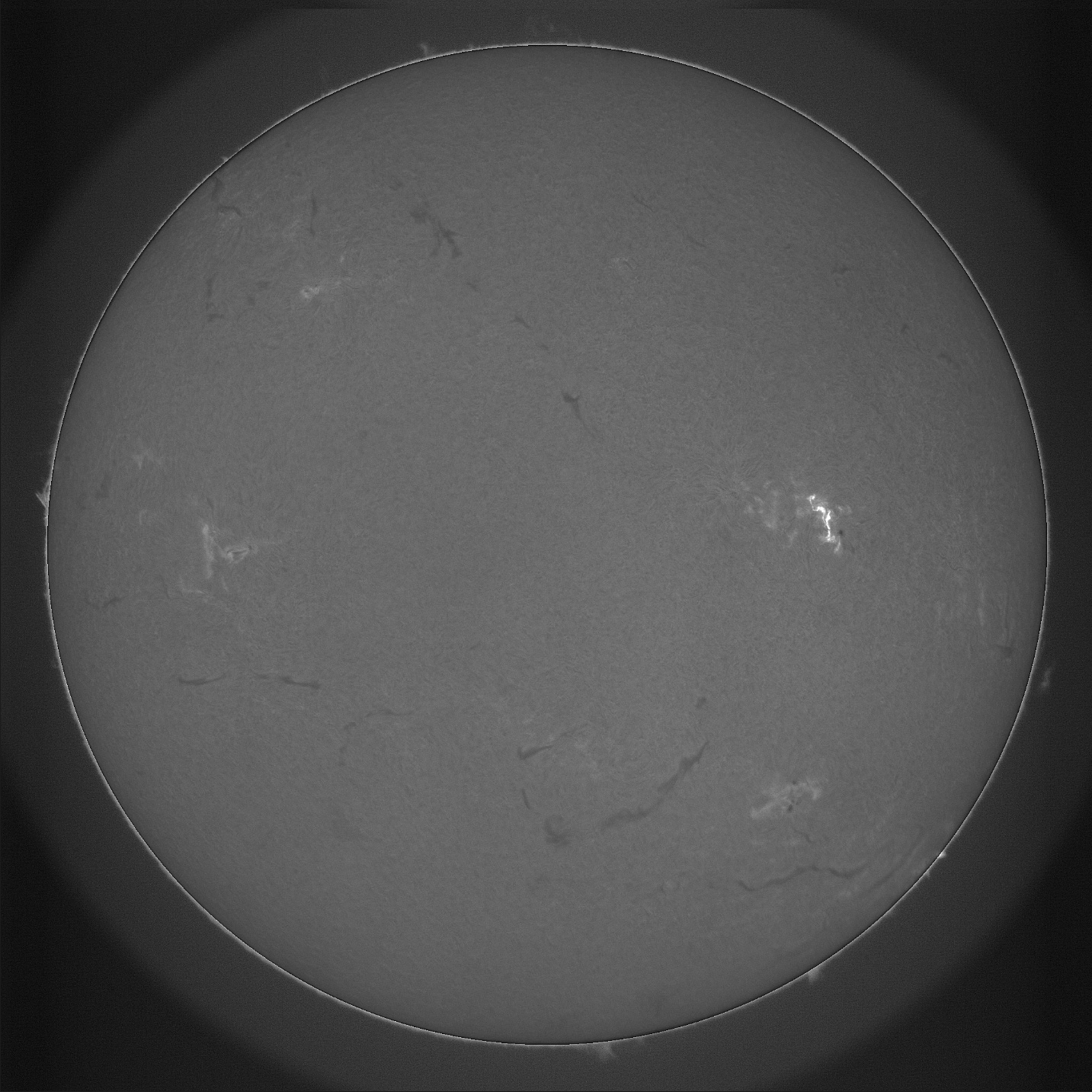}
	}
	\subfigure[08:23:18]{
	\includegraphics[clip=true, trim = 1000px 650px 200px 600px,  width=0.3\linewidth]{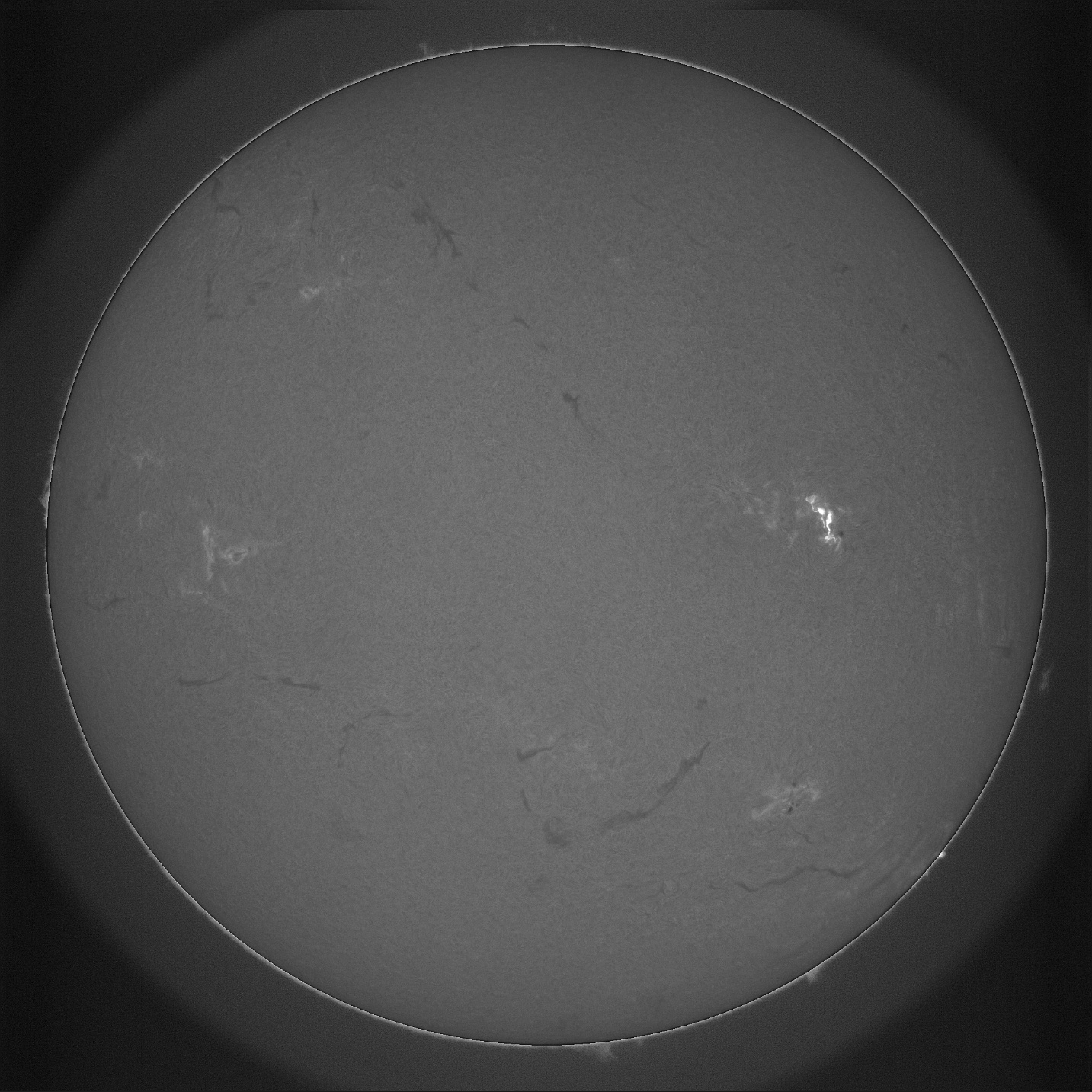}
	}
	\caption{The cropped image sequence shows an occurring flare. The H$\alpha$ images were taken 27.04.2012 by KSO.}
	\label{fig:introFlare}
\end{figure}

\begin{figure}
	\center
	\subfigure[12:50:25]{
	\includegraphics[clip=true, trim = 200px 700px 1000px 500px, width=0.3\textwidth]{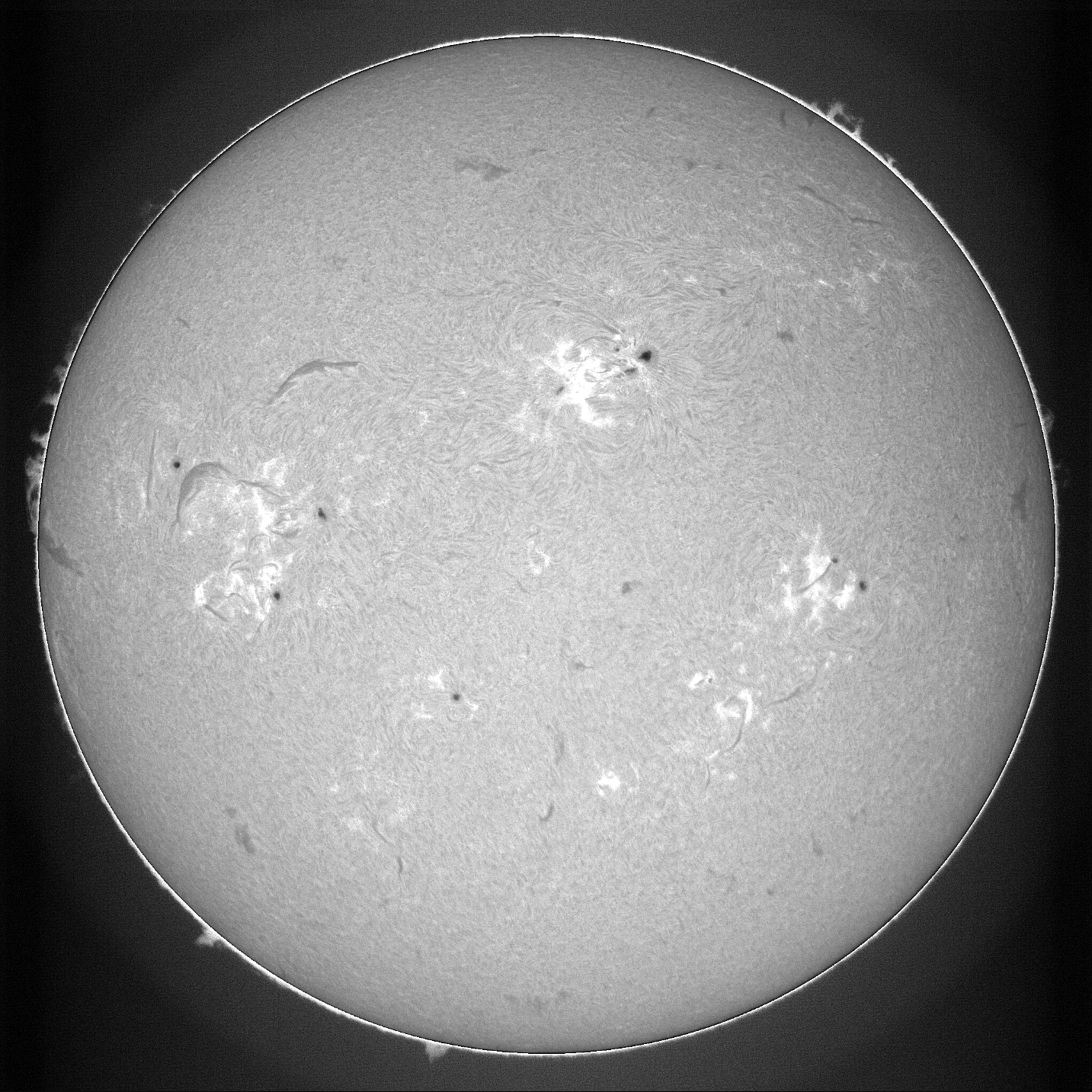}
	}
	\subfigure[13:01:01]{
	\includegraphics[clip=true, trim = 200px 700px 1000px 500px, width=0.3\textwidth] {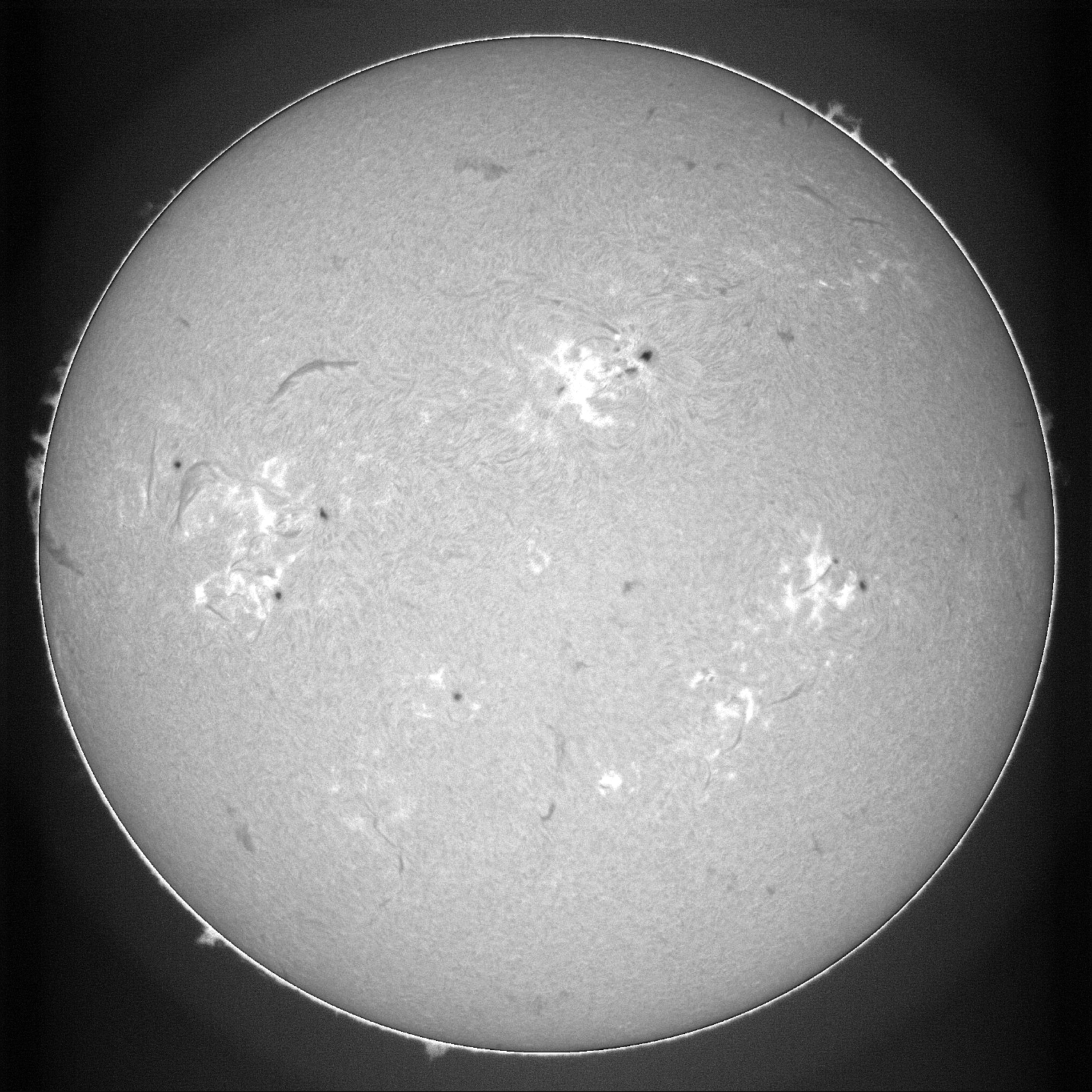}
	}	
	\subfigure[13:04:06]{
	\includegraphics[clip=true, trim = 200px 700px 1000px 500px, width=0.3\textwidth]{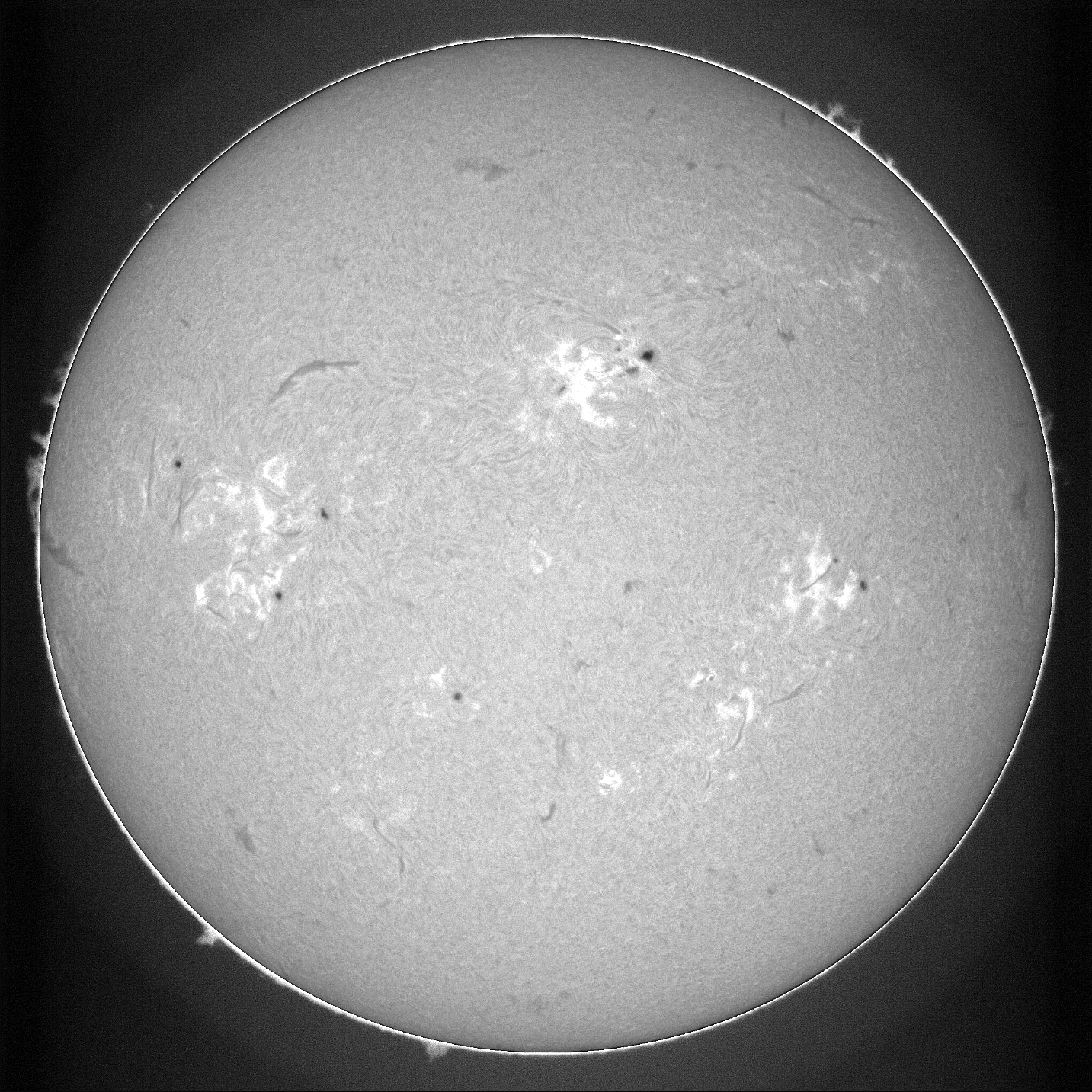}
	}
	\caption{The cropped image sequence shows an erupting filament. The H$\alpha$ images were taken 09.11.2011 by KSO.}
	\label{fig:introFilamenteruption}
\end{figure}

Flares are characterized in H$\alpha$ images by a strong brightness increase of localized sun areas, reaching the maximum extent and intensity within some minutes up to some ten minutes \cite{temmer2001statistical}, followed by a gradual decay of intensity. In figure \ref{fig:introFlare} this event is illustrated. Filaments appear as elongated regions of low intensity on the solar disk of H$\alpha$ images. They have typically about $10\%$ of the disk intensity but are still brighter than the sky \cite{zirin1988astrophysics}. Figure \ref{fig:introFilamenteruption} shows a filament eruption in a H$\alpha$ image sequence.

In the Space Situational Awareness\footnote{\url{http://www.esa.int/Our_Activities/Operations/Space_Situational_Awareness2}, last checked on 06.03.2013} (\emph{SSA}) program of the European Space Agency (\emph{ESA}) the University of Graz is implementing a service to provide near realtime information about the solar activities. This includes the ground-based observation of the sun at the Kanzelhöhe Observatory for Solar and Environmental Research (\emph{KSO}), but also the automatic recognition of solar activities in H$\alpha$ image sequences, which is the aim of this work. The images are taken in a period of ten seconds, if the sight conditions are well, have a resolution of $2048 \times 2048$ pixels and a radiometric resolution of $12$ bits per pixel.

The remainder of this paper is organized as follows. In section \ref{sec:relatedwork} we outline related work dedicated to filament and flare detection in H$\alpha$ images. Our method for the detection and segmentation of these events is presented in section \ref{sec:method}. Section \ref{sec:evalutation} demonstrates our experimental results and section \ref{sec:conclusion} concludes our work.

\section{Related Work} \label{sec:relatedwork}
The work related to image processing in H$\alpha$ images for event detection is divided in flare detection and filament detection. For the flare detection the earliest work was done by \cite{veronig2000automatic}. The segmentation is performed by first selecting key pixels from the images that have twice the mean intensity and using them as an initialization for region growing. This idea was extended by \cite{fernandez2002automatic,qu2003automatic,qu2004automatic} by adding further pre- and postprocessing steps and utilizing machine learning to recognize images that show a flare event.

Filament detection is more difficult and the dedicated work therefore more comprehensive. The first published approaches are also based on selecting key pixels based on local intensities and region growing \cite{aboudarham2008automatic,Gao2002,fuller2004automatic,Fuller2005,Qahwaji2005}. Another common method is based on global and adaptive thresholding \cite{Bernasconi2005,qu2005automatic,Shih2003,Yuan2011}. This methods involve also different preprocessing steps to remove intensity-varieties caused by limb darkening and clouds. \cite{Yuan2011,Bernasconi2005} for example fitted a polynomial and subtracted it from the image to obtain a more homogeneous solar disk. In \cite{Zharkova2003,Zharkova2005} an artificial neural network was trained for the binary classification, if a pixel belongs to a filament or not, where they used the local $3 \times 3$ neighbourhood as input.

\section{Method} \label{sec:method}
The proposed method combines the detection of filaments and flares in one framework by segmenting the H$\alpha$ images. We can divide our method in four main steps. In the preprocssing we normalize the images in terms of intensity and spatial displacement. We further remove noise and local intensity variations with a structural bandpass filter. In the next step we extract and learn features to train a classificator that is able to discriminate between filaments, flares and background. The result of the classification is further regularized by a variational multi-label segmentation approach. In a last postprocessing step we assign an ID to each filament and flare, track them through the sequence and deduce properties like length, area and intensity distribution from them to categorize the events.

\subsection{Preprocessing}
The H$\alpha$ images taken from ground-based observatories can be disturbed in many ways. Variances in the atmosphere have an impact on the exposure time and thus the intensity distribution in the images is shifted and stretched. Additionally the solar disk is not always in the center of the image. Another problem arises from clouds that darken partial regions of the image, which has clearly an impact on the intensity based detection.

To address the first mentioned problem we normalize the intensity of the input image $f$ to a zero mean and a unit standard deviation as follows

\begin{align}
	f_n = \frac{f - \mu}{\sigma}
\end{align}

where $\mu$ denotes to the sample mean and $\sigma$ to the sample standard deviation of the image $f$.

The solar disk is not always in the center of the image and the position can vary in the image sequence. This has influences on the further postprocessing and therefore we register every image in the sequence onto the first image. The displacement vector $\v{u} = (u_1, u_2)^T$ is obtained by minimizing the least squares error and linearisation \cite{lucas1981iterative}. On each level $L$ of a Gaussian image pyramid the displacement vector is given by

\begin{align}
	\v{u}_L = 
	\begin{bmatrix}
		\sum_{i,j} |\nabla f|_x^2 & \sum_{i,j} |\nabla f|_x |\nabla f|_y \\
		\sum_{i,j} |\nabla f|_x |\nabla f|_y & \sum_{i,j} |\nabla f|_y^2
	\end{bmatrix}^{-1}
	\begin{bmatrix}
		\sum_{i, j} |\nabla f_x| (|\nabla f| - |\nabla g|) \\
		\sum_{i, j} |\nabla f_y| (|\nabla f| - |\nabla g|)
	\end{bmatrix}
\end{align}

where $g$ and $f$ denote to the normalized reference image and any normalized further image in the sequence, respectively and $|\nabla f|_x$ is the partial derivative of $|\nabla f|$ in $x$. We use the gradient magnitude $|\nabla f|$ and $|\nabla g|$ to be more robust to image intensity variations.

As a last step in the preprocessing additive noise and large-scale intensity changes, caused by limb darkening and clouds, are removed by applying a structural bandpass filter. The TVL$_1$ model \cite{alliney1992digital,alliney1997property,nikolova2002minimizers,nikolova2004variational} is a variational denoising approach with a total variation (\emph{TV}) regularization and a $L_1$ data fit term.  It is able to remove additive noise while preserving sharp discontinuities and is further almost contrast invariant. The TVL$_1$ model is defined by the following optimization problem

\begin{align}
	\min_u \int_\Omega \textrm{d} |\nabla u| + \lambda \int_\Omega |u - f| \textrm{d}x
\end{align}

where $f$ is the noisy observation, $u$ the solution that minimizes the optimization problem, and $\lambda$ a free parameter that controls the trade-off between the data fidelity (right term) and the total variation regularization (left term).

In the work of Chan and Esedoglu \cite{chan2005aspects} it is shown that the images obtained by changing $\lambda$ generate an interesting scale space. By lowering $\lambda$ objects in the image maintain their contrast with respect to their neighbors, while the boundaries might gradually smooth out. However, if $\lambda$ reaches a critical value, the object suddenly merges with the neighborhood. It is further demonstrated in an example that $\lambda$ is related to the radius $r$ of a disk. In the TVL$_1$ model disks are removed, if the relation $0 \le \lambda <  \frac{2}{r}$ is satisfied.

\begin{figure}
	\center
	\subfigure[$v_1$]{
	\includegraphics[width=0.3\textwidth]{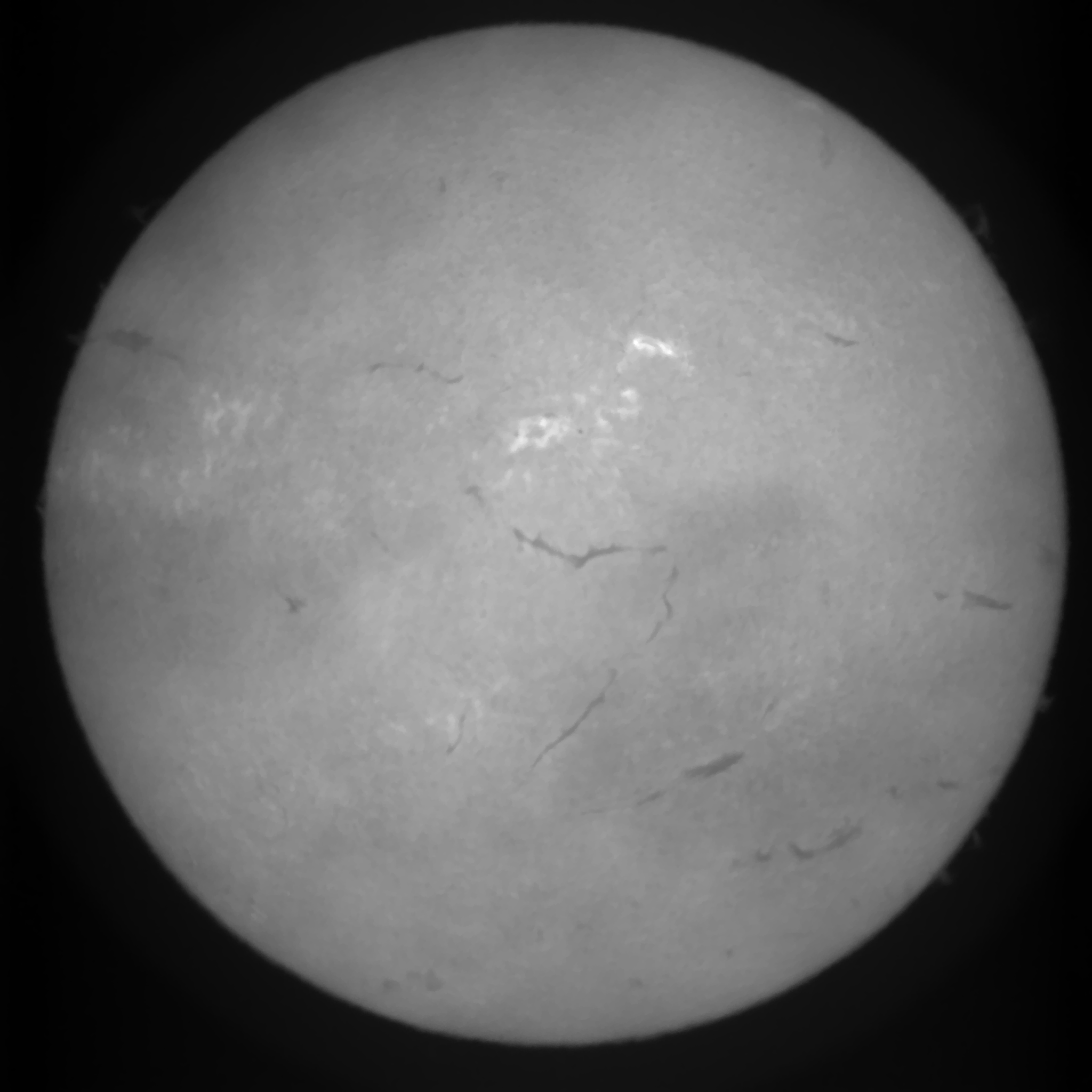}
	}
	\subfigure[$v_2$]{
	\includegraphics[width=0.3\textwidth]{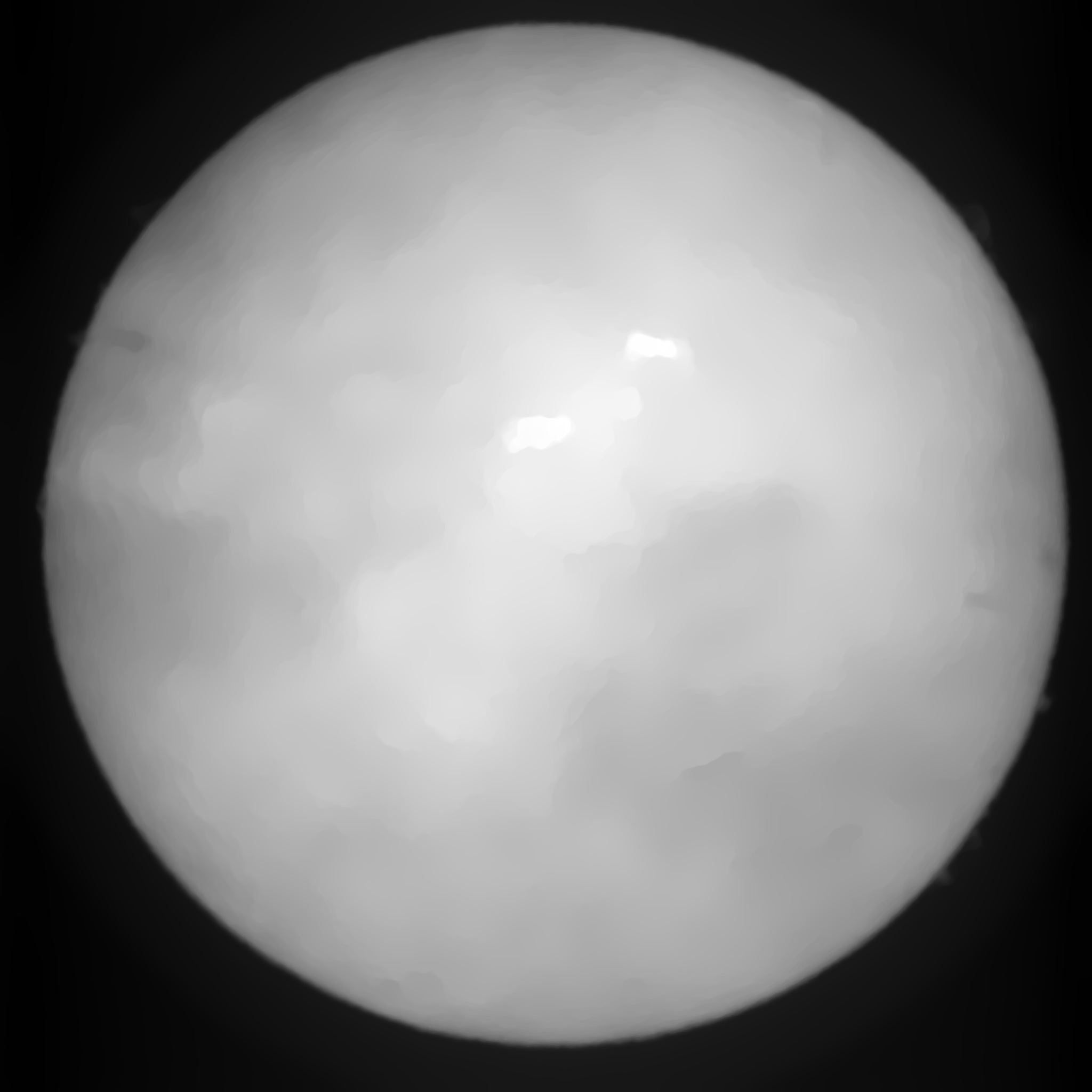}
	}
	\subfigure[$u$]{
	\includegraphics[width=0.3\textwidth]{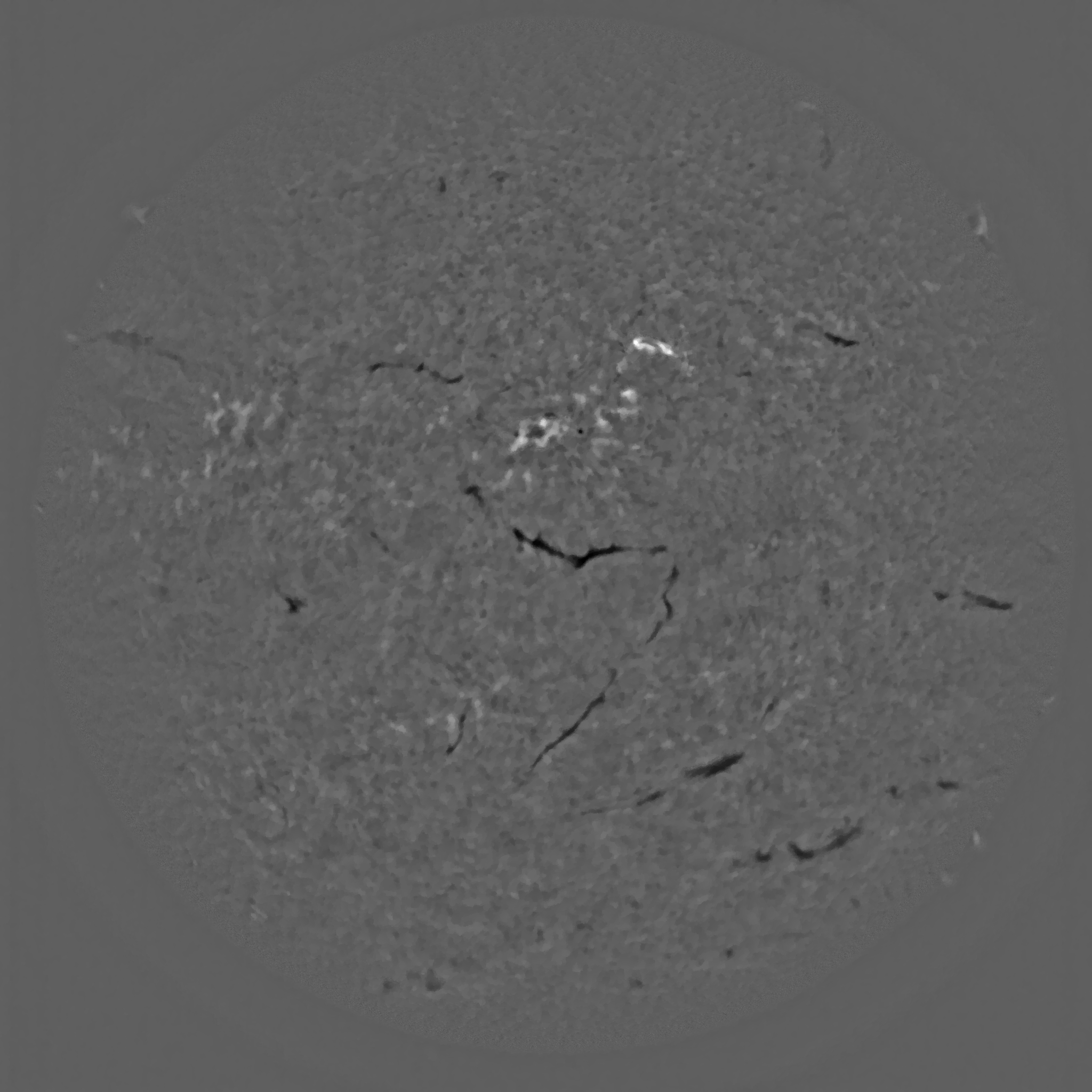}
	}
	\caption{The depicted images show the result of the structural bandpass filter. Figure a illustrates the denoised image $v_1$. The influences of the clouds are clearly visible as darker areas on the left and the right of the solar disk. These large-scale intensity effects are captured in $v_2$, but finer structures like filaments are not captured by this filter. The result $u$ of the structural bandpass filter is shown in figure c.}
	\label{fig:structuralbandpass}
\end{figure}

We utilize this fact in our structural bandpass filter. First, small scaled noise is removed from the image by using $\lambda_1 = 0.9$. In a second round we want to capture the large scaled intensity variations caused by clouds and the limb darkening and use therefore $\lambda_2 = 0.1$. The result of the structural bandpass filter is then given if we subtract the second image $v_2$ from the first image $v_1$. The algorithm is also shown in Algorithm \ref{alg:structuralbandpass} and the results of the single steps are illustrated in figure \ref{fig:structuralbandpass}. To minimize the particular TVL$_1$ models we use the primal-dual algorithm by Chambolle and Pock \cite{chambolle2011first} that can be efficiently implemented on the GPU.

\begin{algorithm}
\caption{Structural Bandpass Filter for an input image $f$}
\label{alg:structuralbandpass}
\begin{algorithmic}
\STATE Choose ${\displaystyle \lambda_1 > \lambda_2 > 0}$
\STATE ${\displaystyle v_1 = \argmin_v \int_\Omega \textrm{d} |\nabla v| + \lambda_1 \int_\Omega |v - f| \textrm{d}x}$
\STATE ${\displaystyle v_2 = \argmin_v \int_\Omega \textrm{d} |\nabla v| + \lambda_2 \int_\Omega |v - f| \textrm{d}x}$
\STATE ${\displaystyle u = v_1 - v_2}$
\end{algorithmic}
\end{algorithm}

\subsection{Feature Selection and Learning}
On the preprocessed images we have to compute and learn descriptive features. An obvious choice is based on the intensity. We created training samples by partial annotating $33$ from different days of the year 2011 and 2012. From the training data, we created histograms for the classes sun spot, filament, flare and background. The estimated distributions have some overlap, especially between filament and background. To further improve the recognition of filaments, we employed the fact that filaments are elongated structures and give strong response to a Hessian filter \cite{frangi1998multiscale}. However, this filter yields also a lot of response in the background, near plagues (bright regions) and on fibrils (dark, small elongated structures). Therefore this feature added almost no additional discriminative information. 

Further, the contrast decreases from the center towards the limb. To incorporate this effect, the distance from the solar disk center to the pixel location has proven to be a valuable feature.


Using the intensity and the distance as features we learned a Gaussian mixture model of the intensity distribution per class using an expectation maximization algorithm~\cite{dempster1977maximum}.

\subsection{Multi-Label Segmentation}
To enforce a smooth segmentation result we minimize the total length of the interface between the classes. This problem is known as the minimal partition problem and can formally be written as

\begin{align}
	\begin{split}
        & \min_{\{ \Omega_i\}_{i=1}^K} \frac{1}{2} \sum_{i=1}^K \text{Per}(\Omega_i) + \lambda \sum_{i=1}^K \int_{\Omega_i} p_i(x) dx \\
        & \text{s.t. } \Omega = \bigcup_{i=1}^K \Omega_i, \quad \Omega_i \cap \Omega_j = \emptyset \;\forall i \neq j 
    \end{split}
\end{align}

where $\Omega$ denotes the image domain, $\text{Per}(\Omega_i)$ the perimiter of the segment $i$ and $p_i$ are pointwise weighting functions, in our case the negative log-probabilities obtained by the Gaussian mixture models of class $i$. To further stabilize the segmentation result, especially for filaments, we compute an exponential weighted average of the probabilities over time. 

To compute the solution, we use a convex relaxation of the given problem \cite{zach2008fast} and minimize the scheme by discretization and applying the primal-dual algorithm according to \cite{chambolle2011first}.

\subsection{Postprocessing}
The most important task in the postprocessing step is the tracking of the flares and filaments in the image sequences. Therefore every connected component in the binary segmentation masks is assigned a unique ID. It is common that flares occur in two ribbons and that a single filament is seen in the H$\alpha$ images as many interrupted pieces. These parts should be considered as one single component and to solve this, we assign the same ID if the distance is less then $25$ px for filaments and $150$ px for flares, respectively. These parameters, as well as the others were determined through numerous experiments.

The IDs for the filaments and flares should also stay the same over time in the image sequence. For that tracking issue we propagate the IDs from previous images. Each component of the actual image collects ID votes from $t$ previous images. The vote count is simply defined by the overlapping pixels for this component. Then the filament derives the ID that gained the most votes from the previous images.

\begin{figure}
	\center
	\subfigure[Input image]{
	\includegraphics[clip=true, trim = 50px 350px 100px 950px, width=\textwidth]{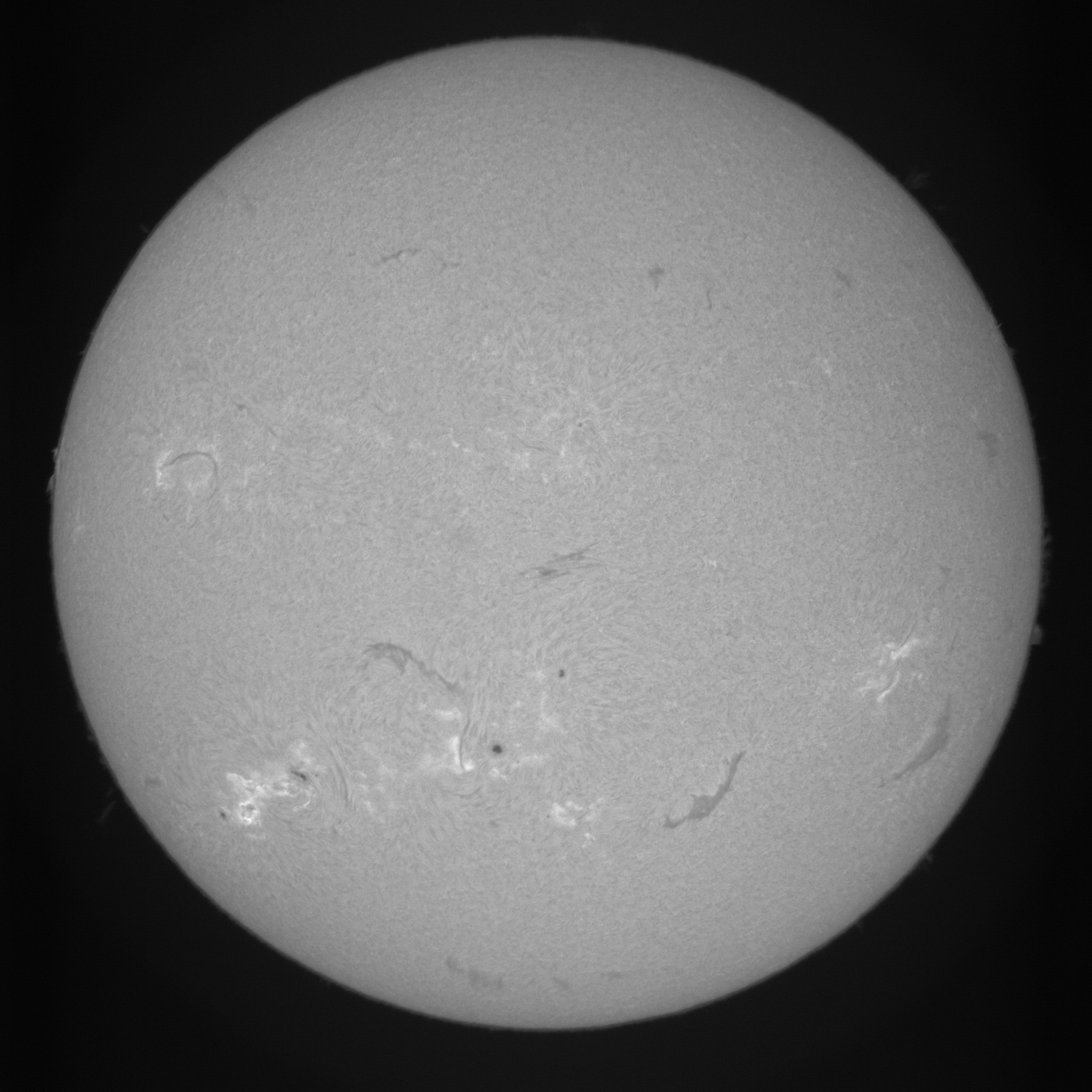}
	}
	\subfigure[Filament detection result]{
	\includegraphics[clip=true, trim = 50px 350px 100px 950px, width=\textwidth]{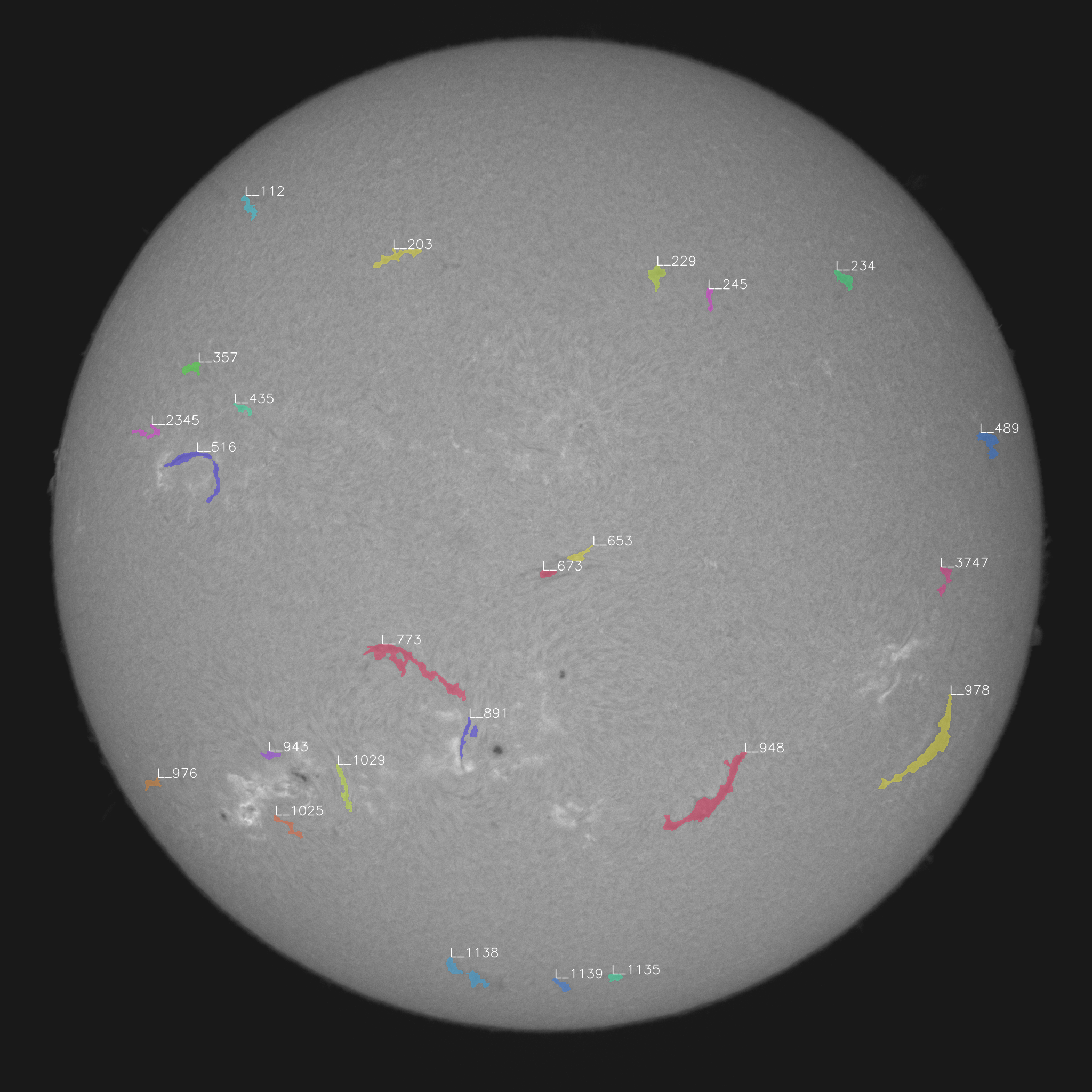}
	}
	\caption{The right image shows the remaining segmented filaments after the postprocessing step, where small components are removed and nearby components are grouped together.}
	\label{fig:output}
\end{figure}

For the filaments two problems remain after the segmentation. The first one is the border of sun spots. While sun spots have usually a lower intensity than filaments the border region is exactly captured by the filament class. Therefore we remove filament components in a radius of $r = 20 \text{px}$ around the sun spot border. A similar problem arises near and in plagues, the brighter regions in the H$\alpha$ image. These filament detections are more difficult to handle, because especially in this region true filaments can exist. We just remove components in this active regions that are not elongated by defining a compactness measurement. The compactness measurement is defined by the area that overlaps with the area of a fitted circle. An example result for the filament detection is given in figure \ref{fig:output}.

The last step is to derive properties from the components for the categorization of flares and filaments. Flares are classified in H$\alpha$ images by the area and the relative intensity to the solar disk \cite{bhatnagar2005fundamentals}. These values can easily be derived from the segmented components. For the filaments the length is an important property. To derive the length of the filament components we perform a thinning operation \cite{guo1989parallel}. From the remaining pixels of the components we can construct a tree  and compute the longest path using the Floyd-Warshall algorithm \cite{floyd1962algorithm}.

\section{Experimental Results} \label{sec:evalutation}
To evaluate the proposed method we compare the automatically detected events to officially registered data by the National Oceanic and Atmospheric Administration (\emph{NOAA})\footnote{\url{http://www.swpc.noaa.gov/ftpmenu/indices/events.html}}, and to the recordings of experts from the KSO, who visually inspected the H$\alpha$ image sequences per day.

For the evaluation of the flare and filament eruption detection, we tested our method on the archived images of July 2012. We selected images in a period of $30$ seconds, independent of the image qualities. 

For the flares, we considered all relevant flares of the category $1$ and higher. This means, the flares exceed a certain area in sun hemispheres \cite{bhatnagar2005fundamentals}. Table \ref{tab:flareresults1+} summarizes the flare detection results in number of true positives, false positives and false negatives. We count a flare detection as true positive, if the start and end time is in the range of $10$ minutes to the official data, the category is correct and location differs not more then $10$ degrees. The obtained results relate to a precision of $1.0$, a recall of $0.85714$ and a $F$-score of $0.92307$.

\begin{table}[ht]\footnotesize
\center
\begin{tabular}{|r|c|c|c|}
\hline
date & true positives & false positives & false negatives \\ \hline
2012-07-02 & 1 & 0 & 0\\ \hline 
2012-07-03 & 3 & 0 & 0\\ \hline 
2012-07-04 & 2 & 0 & 0\\ \hline 
2012-07-05 & 2 & 0 & 0\\ \hline 
2012-07-06 & 3 & 0 & 0\\ \hline 
2012-07-08 & 0 & 0 & 2\\ \hline 
2012-07-09 & 2 & 0 & 0\\ \hline 
2012-07-10 & 2 & 0 & 0\\ \hline 
2012-07-11 & 0 & 0 & 1\\ \hline 
2012-07-14 & 1 & 0 & 0\\ \hline 
2012-07-29 & 1 & 0 & 0\\ \hline 
2012-07-30 & 1 & 0 & 0\\ \hline 
\hline
\textbf{total} & \textbf{18} & \textbf{0} & \textbf{3} \\ \hline
\end{tabular}
\caption{Evaluation of the flare detection of category $1$ and above in H$\alpha$ image sequences. The table presents only days, where a flare with importance class 1, or higher occurred.}
\label{tab:flareresults1+}
\end{table}

For the evaluation of the filament segmentation, we randomly selected one image per day of the month July 2012 and an expert annotated the filaments pixelwise. By comparing the  created ground truth with the results computed by our method we yield an overlap of $0.799825$ by measuring the intersection over union ratio. For precision and recall we obtain the values $0.92985$ and $0.851186$, respectively and the $F$-score is equal to $0.888781$. It should be noted that the filament segmentation is not unique, due to the lack of unique edges and a low contrast to the background.

The last task is the detection of filament eruption. Again, we compared our data to the registered events of the NAOO and the KSO. We defined a filament eruption in our method, if we can not track a filament for $15$ minutes in the image sequence. Table \ref{tab:filamenteruptionsJuly2012} summarizes the evaluation for the month July 2012. Overall, four filament eruptions occurred in July 2012, whereas the method was able to detect all of them correctly. Unfortunately, we detected one false positive on 2012-07-03. In this case the method failed, because the filament was not segmented in several images due to clouds. This implies a precision of $0.80000$, a recall of $1.00000$ and a $F$-score of $0.88889$.

\begin{table}[ht]\footnotesize
\center
\begin{tabular}{|r|c|c|c|}
\hline
date & true positives & false positives & false negatives \\ \hline
2012-07-03 & 0 & 1 & 0 \\ \hline
2012-07-08 & 1 & 0 & 0 \\ \hline
2012-07-11 & 2 & 0 & 0 \\ \hline
2012-07-26 & 1 & 0 & 0 \\ \hline
\hline
\textbf{total} & \textbf{4} & \textbf{1} & \textbf{0} \\ \hline
\end{tabular}

\caption{Evaluation of the filament eruptions in the H$\alpha$ image sequences of July 2013. The table presents only days, where a filament eruption occurred.}
\label{tab:filamenteruptionsJuly2012}
\end{table}

The detection of flares and filament eruptions in the H$\alpha$ images basically share the same problems. The segmentation works fine, until no large clouds cover the solar disk, or the events occur not to close to the limb. In those cases the segmentation might fail, especially for filaments. Further, these cases are one of the main reasons why a good reconnection of the single filament parts is needed.

\section{Conclusions} \label{sec:conclusion}
In this paper we presented a new method that combines the detection of filaments and flares in H$\alpha$ image sequences. We demonstrated a preprocessing chain that performs a normalization, registration, and a structural bandpass filter of the image data. From annotated images we learned Gaussian mixture models for intensity distributions of the sun spots, filaments, flares and background. To regularize the resulting classification, we applied a variational segmentation model. After some additional postprocessing steps we can derive properties that allow a categorization and further a comparison with official data. The results demonstrate that our method is precise in terms of localization and classification.

Future work will evaluate the method continually on realtime data at the KSO. Further, additional research can investigate new features to enhance elongated structures and enhance the detection results of filaments. Further, the grouping of the filament components could be improved, for example by coherence enhancing diffusion or a strategy based on tensor voting.


{\scriptsize \bibliography{refs}}
\end{document}